\pdfoutput=1

\documentclass[11pt]{article}

\usepackage{acl}

\usepackage{times}
\usepackage{latexsym}
\usepackage{color}
\usepackage{xcolor}
\usepackage[T1]{fontenc}

\usepackage[utf8]{inputenc}

\usepackage{microtype}
\usepackage{nert}
\usepackage{tipa}

\title{IruMozhi: Automatically classifying diglossia in Tamil}

\author{Kabilan Prasanna \\
  Academies of Loudoun \\
  \eml{kabilanprasanna@gmail.com} \\\And
  Aryaman Arora \\
  Stanford University \\
  \eml{aryamana@stanford.edu} \\}

\begin{document}
\maketitle
\begin{abstract}
Tamil, a Dravidian language of South Asia, is a highly diglossic language with two very different registers in everyday use: Literary Tamil (preferred in writing and formal communication) and Spoken Tamil (confined to speech and informal media). Spoken Tamil is under-supported in modern NLP systems. In this paper, we release IruMozhi, a human-annotated dataset of parallel text in Literary and Spoken Tamil. We train classifiers on the task of identifying which variety a text belongs to. We use these models to gauge the availability of pretraining data in Spoken Tamil, to audit the composition of existing labelled datasets for Tamil, and to encourage future work on the variety.
\end{abstract}

\section{Introduction}

Diglossia is a linguistic phenomenon wherein a community maintains two (or more) varieties of their language, with the appropriate variety to use depending on the social context \citep{ferguson1959diglossia,ferguson1996epilogue}. Prototypically, diglossia manifests as two varieties: a \textbf{high} variety employed in formal contexts and a \textbf{low} variety employed in informal settings. The high variety tends to be standardised and highly preferred in writing and other formal communication (speeches, news broadcasts, etc.), while the low dialect is confined to speech and other personal communication and subject to regional and stylistic variation. Diglossia is thus a challenge for modern NLP systems---accessible training data on the internet usually overrepresents the high variety, while the average user may prefer using the low variety to interact with NLP systems.

Tamil is one such highly diglossic language primarily spoken in the state of Tamil Nadu in India, and in Sri Lanka and Singapore. Tamil belongs to the Dravidian language family, and is the oldest attested language in this group. Literary Tamil is the standardised (high) variety, continuing a more archaic stage\footnote{Literary Tamil traditionally follows the rules described in the \textit{Na\b{n}\b{n}\={u}l}, a 13th-century grammar by Pava\d{n}anti. However, it has been subject to linguistic change since then by e.g.~the coining of new words.} of the language than the low variety termed Spoken Tamil. Spoken Tamil (or Colloquial Tamil) is subject to dialectal variation by geography and caste, but in India there does exist a widely used and understood (but not officially regulated) Standard Spoken Tamil, based primarily on the dialect of educated non-Brahmin urban residents of central Tamil Nadu \citep{annamalai1980some,schiffman1998standardization,schiffman1999reference,saravanan2009debate}. Both forms of the language coexist in complementary social contexts, and thus practical NLP systems should endeavour to support both.

\begin{table}[]
    \centering
    \small
    \begin{tabular}{ll}
    \toprule
        \textbf{English} & The tail is also white.\\
    \midrule
        \textbf{Literary} & vaalu\colorbox{green!30}{\strut m} vellaiy\colorbox{red!30}{\strut aaga} ulladhu \vspace{5pt}\\
        \textbf{Spoken} (1) & vaalu vellaiy\colorbox{red!30}{\strut e} \colorbox{blue!30}{\strut irukku} \vspace{5pt}\\
        \textbf{Spoken} (2) & vaalu\colorbox{green!30}{\strut m}
        \colorbox{blue!30}{\strut white}-\colorbox{red!30}{\strut ah} \colorbox{blue!30}{\strut irruku} \\
    \specialrule{0.75pt}{3pt}{3pt}
        \textbf{English} & Duryodhana's close friend. \\
    \midrule
        \textbf{Literary} & \colorbox{red!30}{\strut th}uriyodhanan\colorbox{green!30}{\strut in} \colorbox{blue!30}{\strut utra} nanban \vspace{5pt}\\
        \textbf{Spoken} (1) & \colorbox{red!30}{\strut dh}uriyodhanan\colorbox{green!30}{\strut oda} \colorbox{blue!30}{\strut nalla} nanban \vspace{5pt}\\ 
        \textbf{Spoken} (2) & \colorbox{red!30}{\strut dh}uriyodhanan-\colorbox{green!30}{\strut oda} \colorbox{blue!30}{\strut uyir} nanban \\
    \bottomrule
    \end{tabular}
    \caption{Two examples from our parallel corpus of Literary and Spoken Tamil showing morphological (\colorbox{green!30}{\strut 1}), phonological (\colorbox{red!30}{\strut 2}), and lexical differences (\colorbox{blue!30}{\strut 3}).}
    \label{tab:example}
\end{table}

\begin{figure*}
    \centering
    \includegraphics[width=\textwidth]{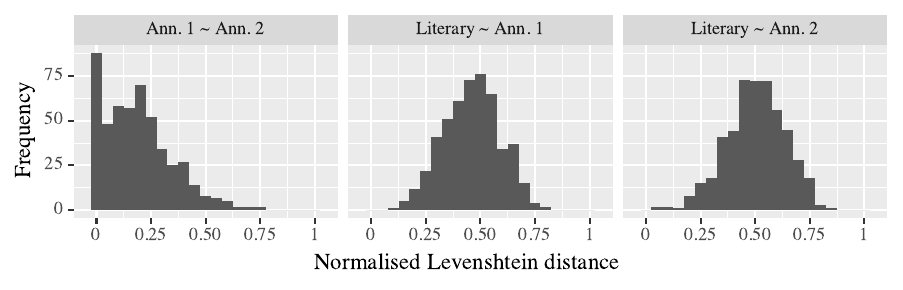}
    \caption{Histogram of normalised Levenshtein distances between parallel sentences from our Literary Tamil corpus and the two Spoken Tamil annotators. The two Spoken Tamil sets are much more similar to each other than to Literary Tamil.}
    \label{fig:levenshtein}
\end{figure*}

Tamil is a rising star in data availability for NLP research \citep{joshi-etal-2020-state,arora-etal-2022-computational}. However, most recent research, particularly on general-purpose systems like language models, has focused on Literary Tamil to the detriment of the Spoken variety. Combined with a lack of standardisation, we expect existing systems to be much worse at all tasks in Spoken Tamil. To combat this problem, we introduce a corpus of high-quality Literary Tamil sentences paired with human-elicited equivalents in Spoken Tamil. Using this data, we train classifiers to identify Spoken Tamil and audit existing Tamil datasets to measure the representation of the two varieties.

\section{Related work}

\paragraph{Spoken Tamil.} While low varieties of diglossic languages are generally understudied in NLP, there is some previous work on NLP for Spoken Tamil. \citet{k-lalitha-devi-2014-automatic} attempted conversion of Spoken Tamil to Literary Tamil using a rule-based system. \citet{nanmalar2022literary,nanmalar2019literary} train models to classify diglossic register for Tamil audio. Finally, recent work on code-switching in Tamil implicitly uses at least some data in Spoken Tamil, since that is the variety most permissive of code-switching \citep{chakravarthi-etal-2020-corpus,chakravarthi-etal-2021-findings-shared,banerjee-etal-2018-dataset,mandl2020overview}.

\paragraph{Diglossia.} Diglossia in NLP has largely been studied in the context of Arabic. For example, \citet{zaidan-callison-burch-2014-arabic,sadat2014automatic,salameh-etal-2018-fine,bouamor-etal-2019-madar} all train models on the task of Arabic dialect and register classification. However, we were inspired to study diglossia in Tamil by \citet{krishna-etal-2022-shot}, the only work on style transfer for Indian languages to our knowledge.

\section{Corpus}

\begin{table}[]
    \centering
    \small
    \begin{tabular}{ll|rrrr}
        \toprule
        \textbf{Set 1} & \textbf{Set 2} & \textbf{Lev.} & \textbf{(norm.)} & \textbf{BLEU} & \textbf{chrF} \\
        \midrule
        Ann.~1 & Ann.~2 & 7.99 & 0.19 & 35.34 & 73.49 \\
        Literary & Ann.~1 & 21.84 & 0.46 & 0.83 & 37.19 \\
        Literary & Ann.~2 & 23.78 & 0.50 & 0.73 & 33.28 \\
        \bottomrule
    \end{tabular}
    \caption{Text similarity metrics between the transliterated Literary Tamil text and the two Spoken Tamil annotators.}
    \label{tab:sim}
\end{table}

\begin{table*}[]
    \centering
    \small
    \begin{tabular}{llllr}
    \toprule
    \textbf{Dataset} & \textbf{Ref.} & \textbf{Register} & \textbf{Source} & \textbf{\# Lines} \\
    \midrule
    IruMozhi & --- & Both & Wikipedia & 1,497 \\
    IruMozhi (augmented) & --- & Both & Wikipedia & 8,634 \\
    \midrule
    Tamilmixsentiment & \citet{chakravarthi-etal-2020-corpus} & Spoken? & YouTube & 15,744 \\
    Offenseval & \citet{chakravarthi-etal-2021-findings-shared} & Spoken? & Social media & 39,527 \\
    Dakshina & \citet{roark-etal-2020-processing} & Literary & Wikipedia & 10,000 \\
    HopeEDI & \citet{chakravarthi-2020-hopeedi} & Spoken? & YouTube & 18,178 \\
    CC-100 & \citet{conneau-etal-2020-unsupervised} & Both? & Web & 6,243,679 \\
    \bottomrule
    \end{tabular}
    \caption{Datasets for romanised Tamil that we consider. The register of each corpus is not known in some cases, in which case we indicate our best guess with `?'.}
    \label{tab:datasets}
\end{table*}

To study diglossia in Tamil, we created IruMozhi,\footnote{\textit{IruMozhi} means `two languages' in Tamil.} a dataset of parallel sentences in Literary and Spoken Tamil. We first collected a high-quality set of 499 sentences randomly sampled from a large corpus of scraped Tamil Wikipedia articles, written in Literary Tamil.\footnote{The articles were scraped in April 2019 and originally hosted as a \href{https://www.kaggle.com/datasets/disisbig/tamil-wikipedia-articles?resource=download}{Kaggle dataset}. We sampled sentences from the first file of the \texttt{train} split of the corpus.} This initial dataset was then converted to Spoken Tamil by two annotators. A few examples of the parallel data are presented in \cref{tab:example}.

\subsection{Annotation}

The dataset from Wikipedia was originally in the Tamil script; however, Spoken Tamil is largely found in the Latin script online. To enable easier comparison to Spoken Tamil and to have parallel romanised training data for both varieties, the dataset was automatically transliterated into the Latin alphabet using a Python program, resulting in the Literary Tamil split of IruMozhi. 

Afterwards, two annotators, both fluent in Literary and Spoken Tamil, were chosen to annotate the sentences into their register of Spoken Tamil. Annotator 1 and 2 both grew up in Salem, Tamil Nadu, India, albeit at different times; annotator 1 tends to use fewer English loanwords.

The annotators were instructed to convert the literary sentences into their register of Tamil while adhering to the original meaning of the sentence as closely as possible. Annotator 1 only had access to the Literary sentences (both Tamil and transliterated), whereas Annotator 2 had access to Annotator 1's conversions as well.

\subsection{Analysis}

\paragraph{Metrics.} We measured Levenshtein distance (raw and normalised), BLEU, and chrF between all three pairings of the transliterated Literary Tamil sentences and the two Spoken Tamil annotated conversions. The latter two metrics were computed using \textsc{SacreBLEU} \citep{post-2018-call}. All metrics are reported in \cref{tab:sim} and \cref{fig:levenshtein}. Overall, the two Spoken Tamil annotators agree with each other more than they do with Literary Tamil across all of our metrics. However, there is clearly linguistic variation in Spoken Tamil given disagreements between the two annotators.

\paragraph{Linguistic features.} We briefly discuss the linguistic differences between Literary and Spoken Tamil. The vowels of Literary Tamil undergo various phonological changes when converted into speech. Vowels, both short monophthongs and diphthongs, are regularly raised in the word-final position. For example, both /-\textipa{a}/ and /-\textipa{aI}/ are raised to [-\textipa{E}]. Word-final /\textipa{u}/ (with the exception of names) is shortened to [\textipa{W}]; additionally, an epenthetic-[\textipa{W}] is usually added to the end of words that end with consonants. When not in the word-final position, /\textipa{e}/ and /\textipa{i}/ are relaxed into /\textipa{E}/ and /\textipa{I}/. Additionally, /\textipa{i}/ along with /\textipa{u}/ are lowered to [\textipa{E}] and [\textipa{o}], respectively, when preceding a short consonant followed by /\textipa{a}/ and /\textipa{aI}/. Unlike the short vowels, long monophthongs will mostly remain the same quality regardless of position.

Word-final nasal consonants (excluding /\textipa{\:n}/) also affect preceding vowels. In all cases, the vowel becomes nasalized and the consonant is dropped. For short vowels, however, the nasal may also change the quality of the vowel. For example, /\textipa{an}/ is nasalized to [\textipa{\~a}], and then raised to [\textipa{\~E}]. Similarly, /\textipa{am}/ is also nasalized to [\textipa{\~a}], but then rounded to [\textipa{\~O}]. 

Outside of regular vowel changes, various other aspects of Spoken Tamil differ from the literary variety. For example, the locative suffix /-\textipa{il}/ is expressed as [-\textipa{lE}]; A suffix like /(-)\textipa{illaI}/, indicating negation, is said as [-\textipa{lE}] at the end of words and [\textipa{illE}] elsewhere. /(-)\textipa{u\:l\:le:}/ (inside) is spoken as [(-)\textipa{u\:l\:lE}]. In some dialects of Spoken Tamil, the 3rd-person irrational ending, /-\textipa{atu}/, can become [-V\textipa{\t{tS}W}] in the past tense of strong verbs, with the vowel depending on the verb being conjugated. In general, strong verbs substitute /-\textipa{tt}-/ and /-\textipa{nt}-/ with [-\textipa{\t{tS}}-] and [-\textipa{n\t{dZ}}-], respectively \citep{schiffman1999reference}.

Finally, there are major lexical differences between Spoken and Literary Tamil. For example, there is a large presence of loanwords in the colloquial form of the language, most often taken from English and Sanskrit. These words, alongside some of native Tamil origin, often replace literary words that may seem too formal in speech. An example of this is \textit{ulladhu}, which is almost always replaced with \textit{irukku} in colloquial contexts as the existence copula. Similarly, the Sanskrit loan \textit{sandosham} is preferred over the native Tamil word \textit{magizhcci} for `happy', although the latter is gaining popularity among the younger generations.


\section{Experiments}

Using IruMozhi, we train models on the task of classifying romanised Tamil text as Literary or Colloquial Tamil. After evaluating our models on a held-out test set, we audit existing datasets of romanised Tamil text to gauge the amount of data available for the two registers.

\begin{table}[]
    \centering
    \small
    \begin{tabular}{llrrrr}
    \toprule
    \textbf{Model} & & \multicolumn{4}{c}{\textbf{Trained on IruMozhi}} \\
    & & Acc. & F1\textsuperscript{ST} & F1\textsuperscript{LT} & Acc.\textsuperscript{D} \\
    \midrule
    Gauss.~NB & $c=4$ & $99.7\%$ & $0.998$ & $0.995$ & $52.9\%$ \\
 & $c=3$ & $99.8\%$ & $0.998$ & $0.996$ & $36.9\%$ \\
 & $c=2$ & $99.8\%$ & $0.998$ & $0.996$ & $58.5\%$ \\
\midrule
    Multi.~NB & $c=4$ & $99.1\%$ & $0.994$ & $0.984$ & $70.8\%$ \\
 & $c=3$ & $98.7\%$ & $0.991$ & $0.978$ & $52.1\%$ \\
 & $c=2$ & $98.8\%$ & $0.992$ & $0.978$ & $20.3\%$ \\
    \midrule
    XLM-R & & $99.4\%$ & $0.996$ & $0.990$ & $81.5\%$ \\
    \bottomrule
    \end{tabular}
    \caption{Results averaged over 5 runs, reporting accuracy and per-class F1 on IruMozhi and accuracy on Dakshina (which the models were not trained on). For all Naïve Bayes models we report with $w=1$.}
    \label{tab:results}
\end{table}

\subsection{Dataset}
We use our parallel corpus of 499 sentences as the training dataset. This gives us 998 sentences in human-annotated Spoken Tamil and 499 sentences in automatically transliterated Literary Tamil. In order to ensure our models do not overfit to a single orthographic standard, we design rules to augment all our data with orthographic variants, resulting in 6,224 Spoken Tamil and 2,410 Literary Tamil sentences. We also strip punctuation and convert all text to lowercase to discourage heuristic overfitting.

One issue is that IruMozhi uses automatically converted Literary Tamil. Fortunately, the Dakshina datatset \citep{roark-etal-2020-processing} contains human-annotated romanised Literary Tamil from the same data distribution as our dataset (Wikipedia). To measure generalisation ability, we check whether models correctly identify Dakshina to be Literary Tamil when only trained on our dataset.

\subsection{Models}
We train two main types of model: \textbf{Naïve Bayes} classifiers on n-gram features and \textbf{XLM-R} finetuned for sequence classification.

For Naïve Bayes, we featurise our data into char and word n-grams using a sliding window, resulting in a fixed-length vector of counts over features for each text input. We test both Gaussian and Multinomial distributions for the feature likelihood, and tune the maximum n-gram length for characters ($c$) and words ($w$) as hyperparameters. We use model implementations from \texttt{scikit-learn}.

XLM-R is a transformer-architecture masked language model trained on the CC-100 web text corpus of one hundred languages, including romanised Tamil \citep{conneau-etal-2020-unsupervised}. Using the HuggingFace implementation of \texttt{XLMRobertaForSequenceClassification}, we train a classification head on the first token \texttt{<s>}. We finetune the entire model for 4 epochs with a learning rate of $2 \cdot 10^{-5}$ for the Adam optimiser.

\subsection{Results and Audits}

\begin{table}[]
    \centering
    \small
    \begin{tabular}{lrr}
    \toprule
    \textbf{Dataset} & \textbf{XLM-R} & \textbf{Multi.~NB} \\
    \midrule
    Tamilmixsentiment & $2.0\%$ & $6.7\%$ \\
    Offenseval & $7.4\%$ & $19.7\%$ \\
    Dakshina & $81.5\%$ & $70.8\%$ \\
    HopeEDI & $6.1\%$ & $20.6\%$ \\
    CC-100 & $23.2\%$ & $13.2\%$ \\
    \bottomrule
    \end{tabular}
    \caption{Estimated percentage of Literary Tamil sentences in each available romanised Tamil corpus, according to finetuned XLM-R and Multinomial Naïve Bayes models trained on IruMozhi.}
    \label{tab:datasets-results}
\end{table}

We present results in \cref{tab:results} (see \cref{sec:more} for results on more hyperparameters). All models reliably converge to near-perfect performance on the held-out portion. However, models vary in their generalisation behaviour; finetuning XLM-R leads to the best performance on Dakshina. Naïve Bayes models, as one would expect, are less reliable for out-of-domain test data.

Having trained these models, we audited the datasets listed in \cref{tab:datasets} to estimate the proportion of Literary and Spoken Tamil in them. We report these estimates in \cref{tab:datasets-results}. Finetuned XLM-R and Multinomial Bayes ($c=4, w=1$) confirm that Dakshina is almost entirely Literary Tamil, while Tamilmixsentiment, Offenseval, and HopeEDI are largely Spoken Tamil. Given the genres that these datasets were collected from (formal Wikipedia vs.~informal social media), these are reasonable estimates. Finally, testing the first 50k lines, we find a surprisingly high portion of Spoken Tamil in the CC-100 \texttt{ta\_rom} split. This suggests that XLM-R was indeed trained on a large amount of Spoken Tamil, explaining why our finetuning was successful.

\section{Conclusion}

We presented IruMozhi, a parallel corpus of Literary and Spoken Tamil annotated on Wikipedia text. We trained models on an augmented version of IruMozhi for classifying Tamil diglossia, and audited the composition of existing datasets and the CC-100 pretraining text in romanised Tamil. We found that there are indeed labelled and unlabelled data sources for Spoken Tamil text, indicating hopeful avenues for future NLP research on the variety.

We hope to train style transfer models for the two varieties and study diglossia in other Indian languages. Our aim is to encourage work on lesser-studied languages and dialects in South Asia.

\bibliography{anthology,custom}
\bibliographystyle{acl_natbib}

\appendix
\onecolumn
\section{More results}
\label{sec:more}

\begin{table*}[h]
    \centering
    \small
    \begin{tabular}{ll|rrrr|rrr}
    \toprule
    \textbf{Model} & \textbf{Params} & \multicolumn{4}{c|}{\textbf{Trained on IruMozhi}} & \multicolumn{3}{c}{\textbf{Trained on IruMozhi + Dakshina}} \\
    & & Acc. & F1\textsuperscript{ST} & F1\textsuperscript{LT} & Acc.\textsuperscript{Dakshina} & Acc. & F1\textsuperscript{ST} & F1\textsuperscript{LT} \\
    \midrule
    Naïve Bayes (Gaussian) & $c=4, w=1$ & $99.7\%$ & $0.998$ & $0.995$ & $52.9\%$ & $99.7\%$ & $0.998$ & $0.995$\\
 & $c=3, w=1$ & $99.8\%$ & $0.998$ & $0.996$ & $36.9\%$ & $99.7\%$ & $0.998$ & $0.994$\\
 & $c=2, w=1$ & $99.8\%$ & $0.998$ & $0.996$ & $58.5\%$ & $99.8\%$ & $0.999$ & $0.997$\\
 & $c=1, w=1$ & $99.4\%$ & $0.996$ & $0.989$ & $91.3\%$ & $99.2\%$ & $0.995$ & $0.987$\\
 & $c=0, w=1$ & $99.4\%$ & $0.996$ & $0.990$ & $1.7\%$ & $99.4\%$ & $0.996$ & $0.988$\\
 & $c=4, w=0$ & $99.1\%$ & $0.994$ & $0.984$ & $48.7\%$ & $99.5\%$ & $0.996$ & $0.991$\\
 & $c=3, w=0$ & $94.3\%$ & $0.959$ & $0.906$ & $29.5\%$ & $93.9\%$ & $0.956$ & $0.901$\\
 & $c=2, w=0$ & $67.3\%$ & $0.708$ & $0.628$ & $43.2\%$ & $68.2\%$ & $0.718$ & $0.636$\\
 & $c=1, w=0$ & $72.2\%$ & $0.761$ & $0.561$ & $29.9\%$ & $40.8\%$ & $0.330$ & $0.470$\\
\midrule
Naïve Bayes (Multinomial) & $c=4, w=1$ & $99.1\%$ & $0.994$ & $0.984$ & $70.8\%$ & $99.1\%$ & $0.994$ & $0.984$\\
 & $c=3, w=1$ & $98.7\%$ & $0.991$ & $0.978$ & $52.1\%$ & $98.4\%$ & $0.989$ & $0.971$\\
 & $c=2, w=1$ & $98.8\%$ & $0.992$ & $0.978$ & $20.3\%$ & $99.0\%$ & $0.993$ & $0.981$\\
 & $c=1, w=1$ & $99.1\%$ & $0.993$ & $0.983$ & $2.2\%$ & $99.0\%$ & $0.993$ & $0.981$\\
 & $c=0, w=1$ & $99.0\%$ & $0.993$ & $0.982$ & $74.0\%$ & $98.4\%$ & $0.989$ & $0.972$\\
 & $c=4, w=0$ & $98.7\%$ & $0.991$ & $0.977$ & $76.0\%$ & $98.1\%$ & $0.987$ & $0.966$\\
 & $c=3, w=0$ & $98.0\%$ & $0.986$ & $0.965$ & $65.4\%$ & $98.6\%$ & $0.990$ & $0.974$\\
 & $c=2, w=0$ & $94.3\%$ & $0.960$ & $0.902$ & $50.9\%$ & $94.2\%$ & $0.959$ & $0.901$\\
 & $c=1, w=0$ & $82.0\%$ & $0.880$ & $0.643$ & $34.8\%$ & $82.6\%$ & $0.884$ & $0.655$\\
    \midrule
    XLM-R & & $99.4\%$ & $0.996$ & $0.990$ & $81.5\%$ & $99.1\%$ & $0.990$ & $0.991$ \\
    \bottomrule
    \end{tabular}
    \caption{Results on more hyperparameter settings.}
    \label{tab:results}
\end{table*}


\end{document}